\documentclass{article}
\usepackage{amsmath,graphicx,mlspconf}
\usepackage{balance}

\copyrightnotice{Accepted to the 2015 IEEE International Workshop on Machine Learning for Signal Processing}
\title{Accelerated graph-based spectral polynomial filters}
\name{Andrew Knyazev and Alexander Malyshev
}
\address{Mitsubishi Electric Research Labs (MERL),
201 Broadway, 8th floor, Cambridge, MA 02139, USA.\\
email: knyazev@merl.com, malyshev@merl.com}
\begin{document}
%\ninept
%

\maketitle
\begin{abstract}
Graph-based spectral denoising is a low-pass filtering using the eigendecomposition
of the graph Laplacian matrix of a noisy signal. Polynomial filtering avoids costly computation
of the eigendecomposition by projections onto suitable Krylov subspaces.
Polynomial filters can be based, e.g.,\ on the bilateral and guided filters. 
We propose constructing accelerated polynomial filters by running flexible Krylov subspace
based linear and eigenvalue solvers such as the Block Locally Optimal Preconditioned
Conjugate Gradient (LOBPCG) method.
\end{abstract}
\section{Introduction}
\label{sec:intro}

In this note, we deal with noise removal from a given noisy signal, which is a basic problem
in signal processing, with applications, e.g.,\ in image denoising \cite{M13}.
Apart from the trivial application of removing noise prior to presenting
the image to a human observer, pre-smoothing an image and noise removal may help to improve
performance of many image-processing algorithms, such as compression, enhancement, segmentation etc.
A noise removal operation is often referred to as a filter.

Modern denoising algorithms endeavor to preserve the image details while removing the noise.
One of the most popular denoising filters is the bilateral filter (BF), which
smooths images while preserving edges, by taking the weighted average of the nearby pixels.
The weights depend on both the spatial distance and photometric similarity, thus providing
local adaptivity to the input image. Bilateral filtering was introduced in
\cite{AW95,SB97,TM98} as an intuitive tool
without theoretical justification. Since then, connections between the BF and other well-known
filtering techniques such as anisotropic diffusion, weighted least squares, Bayesian methods,
kernel regression and non-local means have been explored, see, e.g., survey \cite{PKTD09}.

A convenient way to represent images is by graphs \cite{SNFOV13}, especially when considering
images over irregular grids and treating non-local interactions between pixels.
A single application of the bilateral filter to an image may be interpreted
as a vertex domain transform on a graph with pixels as vertices,
intensity values of the pixels as the graph signal and filter coefficients
as link weights that capture the similarity between the vertices. The BF transform is a nonlinear
anisotropic diffusion, cf. \cite{PM87,PM90,D02}, determined by entries of the graph Laplacian matrix,
which are related to the BF weights. In the linear case, the solutions of time-dependent
anisotropic diffusion problems are represented by operator exponentials or semigroups, which can be
approximated by operator polynomials.

Eigenvalues and eigenvectors of the graph Laplacian matrix allow us to apply the Fourier analysis
to the graph signals or images as in \cite{SNFOV13} and perform frequency selective
filtering operations on graphs, similar to those in traditional signal processing.
Expensive computation of eigenvectors is avoided in polynomial spectral filters,
which are fully implemented in the vertex domain. For example, the spectral filters
in \cite{GNO13,TKMV14} are based on optimal polynomials constructed by means of the Chebyshev approximation and conjugate gradient algorithm.

A recent newcomer in the field of filtering is the guided filter (GF) proposed in
\cite{HST13,HS15} and included into the image processing toolbox of MATLAB.
According to our limited experience, the GF filter is significantly faster
than the BF filter. The authors of \cite{HST13} additionally advocate that
GF is gradient preserving and avoids the gradient reversal problems
in contrast to BF, which is not gradient preserving.

Iterative application of smoothing filters like BF and GF can be interpreted
as matrix power transforms, in general case, nonlinear, or, equivalently as
explicit integration in time of the corresponding anisotropic diffusion equation.
In the present paper, we continue the approach of \cite{GNO13,TKMV14} and propose
acceleration of the iterative smoothing filters based on the polynomial approximations
implicitly constructed in the conjugate gradient (CG) algorithm and in the LOBPCG,
which is a leading eigensolver for large symmetric matrices \cite{K01}.
LOBPCG has been successfully applied to image segmentation in \cite{K03}.

The remainder of the paper is organized as follows. Section 2 gives main formulas
of the original bilateral filter in the graph-based notation. Section 3 provides
a similar description of the guided filter from \cite{HST13}. Section 4 introduces
a basic Fourier calculus on graphs and applies it to a simplified frequency analysis
of iterations with the graph Laplacian matrix.
Section 5 describes the CG acceleration of the smoothing filters.
Section 6 gives a brief description of the LOBPCG method and suggests it as an acceleration
for BF and GF. Our numerical experiments in Section 7 demonstrate improved performance
of our accelerated filters in the one-dimensional case mostly for the BF filter.
Similar improvements are expected when accelerating the GF filter.

\section{Bilateral filter as a vector transform}
\label{sec:BF}

The bilateral filter transforms an input image $x$ into the output image $y$
by the weighted average of the pixels of $x$:
\begin{equation}\label{eq1}
 y_i = \sum_j\frac{w_{ij}}{\sum_jw_{ij}}x_j.
\end{equation}
Let $p_i$ denote the geometrical position of a pixel $i$. Then
\begin{equation}\label{eq2}
 w_{ij}=\exp\left(-\frac{\|p_i-p_j\|^2}{2\sigma_s^2}\right)
 \exp\left(-\frac{|x_i-x_j|^2}{2\sigma_r^2}\right),
\end{equation}
where $\sigma_s$ and $\sigma_r$ are the filter parameters \cite{TM98},
$\|p_i-p_j\|$ is a spatial distance between pixels $i$ and $j$.
For color images, the photometric distance $|x_i-x_j|$
can be computed in the CIE-Lab color space as suggested in \cite{TM98}.

The BF weights $w_{ij}$ determine an undirected graph $G=(\mathcal{V},\mathcal{E})$,
where the vertices $\mathcal{V}=\{1,2,\ldots,N\}$ are the pixels of the input image
and the edges $\mathcal{E}=\{(i,j)\}$ connect the ``neighboring'' pixels $i$ and $j$.
The adjacency matrix $W$ of the graph $G$ is symmetric and has the entries $w_{ij}\ge0$.
Let $D$ be the diagonal matrix with the nonnegative diagonal entries $d_i=\sum_{j}w_{ij}$.
In the matrix notation, the bilateral filter operation (\ref{eq1}) is the vector transform
with $W$ depending on a guidance image (we use $x$ as a guidance image in (\ref{eq2}))
\begin{equation}\label{eq3}
y = D^{-1}Wx=x-D^{-1}Lx,
\end{equation}
where
\begin{equation}\label{eq4}
 L=D-W.
\end{equation}
is the Laplacian matrix for the graph $G$. Gershgorin's theorem from matrix analysis
guarantees a smoothing effect of the bilateral filter, when $\max_id_i\leq2$.

The bilateral filtering can be used iteratively. There are two ways to iterate BF:
(1) by changing the weights $w_{ij}$ at each iteration using the result
of the previous iteration as a guidance image, or (2) by using the fixed weights at each iteration
as calculated from the initial image as a guidance image for all iterations.
The first alternative results in a nonlinear filter, where the BF graph changes at every iteration
and we can only provide separate spectral interpretation for each stage.
The second alternative generates a linear filter, the Laplacian matrix is fixed for all iterations,
and it is possible to provide a spectral interpretation of the whole cascaded operation.
The second way is also faster to compute since the BF weights are computed only once in the beginning.

The fastest implementations of BF have the arithmetical complexity $O(N)$,
where $N$ is the number of pixels or voxels in the input image \cite{YTA09,C13,SK15}.

\section{Guided filter as a vector transform}
\label{sec:GF}

\begin{tabular}{l}
\hline\\[-2ex]
\textbf{Algorithm 1}{ Guided Filter}\\\hline\\[-2ex]
\textbf{Input:} $x$, $g$, $\rho$, $\epsilon$\\
\textbf{Output:} $y$\\
\quad $mean_g=f_{mean}(g,\rho)$\\
\quad $mean_x=f_{mean}(x,\rho)$\\
\quad $corr_g=f_{mean}(g.*g,\rho)$\\
\quad $corr_{gx}=f_{mean}(g.*x,\rho)$\\
\quad $var_g=corr_g-mean_g.*mean_g$\\
\quad $cov_{gx}=corr_{gx}-mean_{g}.*mean_x$\\
\quad $a = cov_{gx}./(var_g+\epsilon)$\\
\quad $b = mean_x-a.*mean_g$\\
\quad $mean_a=f_{mean}(a,\rho)$\\
\quad $mean_b=f_{mean}(b,\rho)$\\
\quad $y=mean_a.*g+mean_b$\\
\\
\end{tabular}

Algorithm 1 is a pseudo-code of the guided filter from \cite{HST13},
where $f_{mean}(\cdot,\rho)$ denotes a mean filter with the window width $\rho$.
The constant $\epsilon$ determines the smoothness degree of the filter: the larger $\epsilon$
the larger a smoothing effect. The arithmetical operations $.*$ and $./$ are the componentwise multiplication and division.

The input image is $x$, the output image is $y$. The guidance image $g$ has the same size as $x$.
If $g$ coincides with $x$, then the guided filter is called self-guided.
The arithmetical complexity of GF equals $O(N)$, where $N$ is the number of pixels in $x$.

The guided filter operation of Algorithm 1 is the transform
\begin{equation}\label{eq5}
y = W(g)x,
\end{equation}
where the implicitly constructed symmetric matrix $W(g)$ has the following entries, see \cite{HST13}: 
\begin{equation}\label{eq6}
W_{ij}(g)=\frac{1}{|\omega|^2}\sum_{k\colon (i,j)\in\omega_k}
\left(1+\frac{(g_i-\mu_k)(g_j-\mu_k)}{\sigma_k^2+\epsilon}\right).
\end{equation}
Here $\omega_k$ is the neighborhood around pixel $k$ of width $\rho$, where
the mean filter $f_{mean}(\cdot,\rho)$ is applied, and $|\omega|$ denotes the number of
pixels in $\omega_k$, the same for all $k$. The values $\mu_k$ and $\sigma_k^2$ are the mean
and variance of the image $g$ in $\omega_k$.

The graph Laplacian matrix is obtained from the matrix $W$ in the standard way.
According to \cite{HST13}, the values $d_i=\sum_{j}w_{ij}$ equal 1. Therefore,
the diagonal matrix $D$ equals the identity matrix $I$, and the graph Laplacian matrix is
the symmetric nonnegative definite matrix $L=I-W$.

The guided filtering can also be used iteratively. Similar to the BF filter,
the guided filter is applied iteratively: (1) either using the result of the previous
iteration as a guidance image, or (2) by using the initial image as a guidance image
for all iterations. The first alternative results in a nonlinear GF filter.
The second alternative generates a linear GF filter, and the Laplacian matrix remains fixed
at each iteration.

\section{Spectral interpretation of the low-pass filters}
\label{sec:spectral}

The graph eigenstructure is the eigenvalues and eigenvectors of the graph
Laplacian matrix~$L$, which is symmetric and nonnegative definite.
In certain situations, the normalized Laplacian matrix
$(\mbox{diag\,}L)^{-1/2}L(\mbox{diag\,}L)^{-1/2}$
may be more suitable than $L$.
The spectral factorization of $L$ is the matrix decomposition
\begin{equation}\label{eq8}
 L = U\Lambda U^T,
\end{equation}
where the diagonal elements $\lambda_i$ of the diagonal matrix $\Lambda$ and
columns $u_i$ of the orthogonal matrix $U=[u_1,\ldots,u_n]$ are, respectively,
the eigenvalues of $L$ and corresponding eigenvectors of the unit 2-norm.

Similar to the classical Fourier transform, the eigenvectors and eigenvalues
of the graph Laplacian matrix $L$ provide the oscillatory structure of graph signals.
The eigenvalues $\lambda_1\leq\cdots\leq\lambda_i\leq\cdots\leq\lambda_n$
can be treated as graph frequencies.
The corresponding eigenvectors $u_i$ of the Laplacian matrix $L$ are generalized
eigenmodes and demonstrate increasing oscillatory behavior as the magnitude
of the graph frequency increases.
The Graph Fourier Transform (GFT) of an image $x$ is defined by the matrix transform
$\widehat{x}=U^Tx$, the inverse GFT is the transform $x=U\widehat{x}$.

Let us consider the BF vector transform (\ref{eq3}). In numerical analysis,
the linear transformations $(D^{-1}W)^k$ are called the power iterations with the
amplification matrix $D^{-1}W$ or simple iterations for the equation $Lx=0$
with the preconditioner $D$. Application of the transform (\ref{eq3}) preserves
the low frequency components of $x$ and attenuates the high frequency components,
cf. \cite{GNO13,TKMV14}. It is also well-known that the Krylov subspaces well
approximate the eigenvectors corresponding to the extreme eigenvalues. Thus the
projections onto the suitable Krylov subspaces would be an appropriate choice
for high- and low-pass filters. The Krylov subspace methods are efficient
owing to their low cost, reasonably good convergence and simple implementation
without painful parameter tuning. The convergence can be accelerated by the aid
of good preconditioners.

\section{CG acceleration}
\label{sec:bfcg}

Since the graph Laplacian matrix $L$ is symmetric and nonnegative definite,
the first candidate to replace the simple iteration $x_{k+1}=x_k-D^{-1}Lx_k$
is the preconditioned conjugate gradient (CG) method for the system of
homogeneous linear equations $Lx=0$ with the preconditioner matrix $D$.
To avoid over-smoothing, only few iterations of the preconditioned CG method
must be performed. A large number of iterations for $Lx=0$ produces piecewise
constant images and is better applicable to the image segmentation problems.
Good references for the Krylov subspace methods are the books \cite{G97,V03}.

\begin{tabular}{l}
\hline\\[-2ex]
\textbf{Algorithm 2}{ Preconditioned CG($k_{\max}$) for $Lx=0$}\\\hline\\[-2ex]
\textbf{Input:} $L$, $x_0$, $k_{\max}$, preconditioner $D$\\
\textbf{Output:} $x$\\
\quad $x=x_0$\\
\quad $r=-Lx$\\
\quad\textbf{for }{$k=1,\ldots,k_{\max}$}\textbf{ do}\\
\qquad $s=D^{-1}r$\\
\qquad\textbf{if }$k=1$\textbf{ then}\\
\qquad\quad $p=s$\\
\qquad\textbf{else}\\
\qquad\quad $\beta=(s^T(r-r_{old}))/(s_{old}^Ts_{old})$\\
\qquad\quad $p=s+\beta p$\\
\qquad\textbf{endif}\\
\qquad $q=Lp$\\
\qquad $\alpha=(s^Tr)/(p^Tq)$\\
\qquad $x=x+\alpha p$\\
\qquad $r_{old}=r$\\
\qquad $s_{old}=s$\\
\qquad $r=r-\alpha q$\\
\quad\textbf{endfor}\\
\end{tabular}

Algorithm 2 is a slightly modified standard preconditioned CG algorithm
formally applied to the system of linear equations $Lx=0$. It contains the
formula for $\beta$ that is different from that in the MATLAB
implementation of the preconditioned CG. Such a formula converts the algorithm
into a flexible variant, which possesses better convergence properties, when
the input matrices $L$ and $D$ may change; see, e.g., \cite{KL08,BDK12}.

The CG iterations are the polynomial filters, i.e., represented in the form
$x_k=p_k(L)x_0$, where the coefficients of the polynomial $p_k(\lambda)$
of degree $k$ depend on the matrix $L$ and initial vector $x_0$.
In the preconditioned case, the filter has the form $x_k=p_k(D^{-1}L)x_0$,
and the coefficients of the polynomial $p_k(\lambda)$ depend on $D^{-1}L$ and $x_0$.

\section{LOBPCG acceleration}
\label{sec:lobpcg}

A more powerful alternative to the Krylov subspace solvers for linear systems as
the polynomial spectral filters are the eigensolvers with preconditioning. 
Since the graph Laplacian matrix $L$ is symmetric, we propose the symmetric eigensolver
LOBPCG with preconditioning for construction of low-pass filters.
LOBPCG is a very efficient eigensolver and gives a fast solution
to the spectral image segmentation problem, see \cite{K03}.
To avoid over-smoothing during noise removal, LOBPCG must perform only few iterations.

Algorithm 3 below is a standard non-blocked version of the preconditioned LOBPCG algorithm,
which smooths the input signal $x_{0}$ with respect to the eigenmodes of the Laplacian
matrix $L$. If necessary, Algorithm 3 can be modified to obey the constraint that
the vectors $x_k$ must be orthogonal to the vector $e$ with all components equal to 1.
The vector $e$ is the eigenmode corresponding to the zero eigenvalue of $L$.

\hspace{-1.5em}
\begin{tabular}{l}
\hline\\[-2ex]
\textbf{Algorithm 3}{ LOBPCG method as a low-pass filter}\\\hline\\[-2ex]
\textbf{Input:} $L$, $D$, $x_0$, \textit{ and a preconditioner }$T$\\
\textbf{Output:} $x_{kmax}$\\
\quad $p_0=0$\\
\quad\textbf{for }{$k=0,\ldots,k_{\max}-1$}\textbf{ do}\\
\qquad $\lambda_k=(x_k^TLx_k)/(x_k^TDx_k)$\\
\qquad $r=Lx_k-\lambda_kDx_k$\\
\qquad $w_k=Tr$\\
\qquad\textit{use the Rayleigh-Ritz method for the pencil $L-\lambda D$}\\
\qquad\quad\textit{on the trial subspace} $span\{w_k,x_k,p_k\}$\\
\qquad $x_{k+1}=w_k+\tau_kx_k+\gamma_kp_k$\\
\qquad\quad\textit{(the Ritz vector for the minimum Ritz value)}\\
\qquad $p_{k+1}=w_k+\gamma_kp_k$\\
\quad\textbf{endfor}\\
\end{tabular}

\section{Numerical experiments with the Krylov subspace-based
polynomial acceleration of the filters}

\begin{figure}[htb]
\begin{minipage}[b]{1.0\linewidth}
  \centering
  \centerline{\includegraphics[width=\columnwidth]{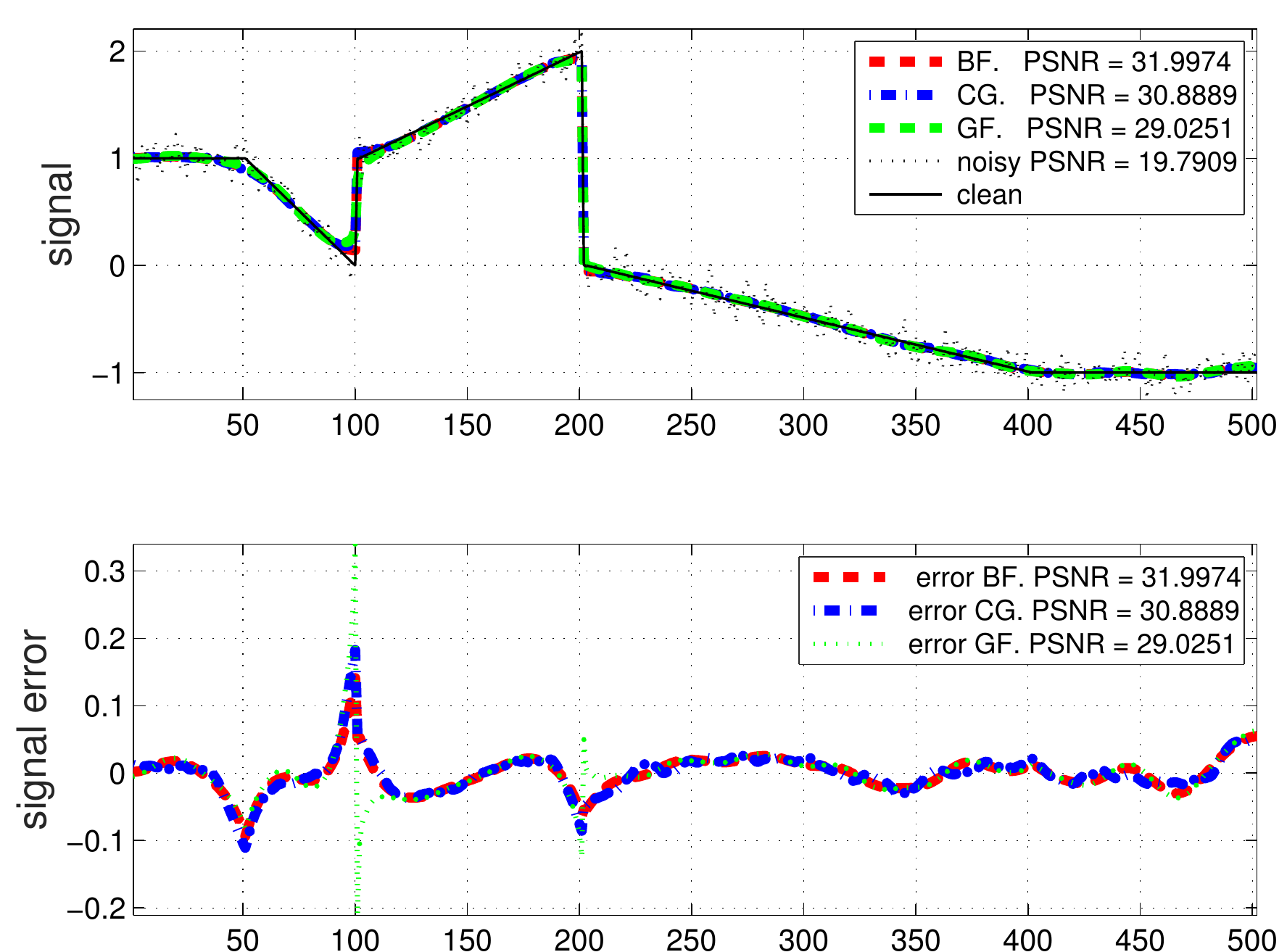}}
\end{minipage}
\caption{Filters in 1D case: 500 points}
\label{fig:1D500}
\end{figure}

\begin{figure}[htb]
\begin{minipage}[b]{1.0\linewidth}
  \centering
  \centerline{\includegraphics[width=\columnwidth]{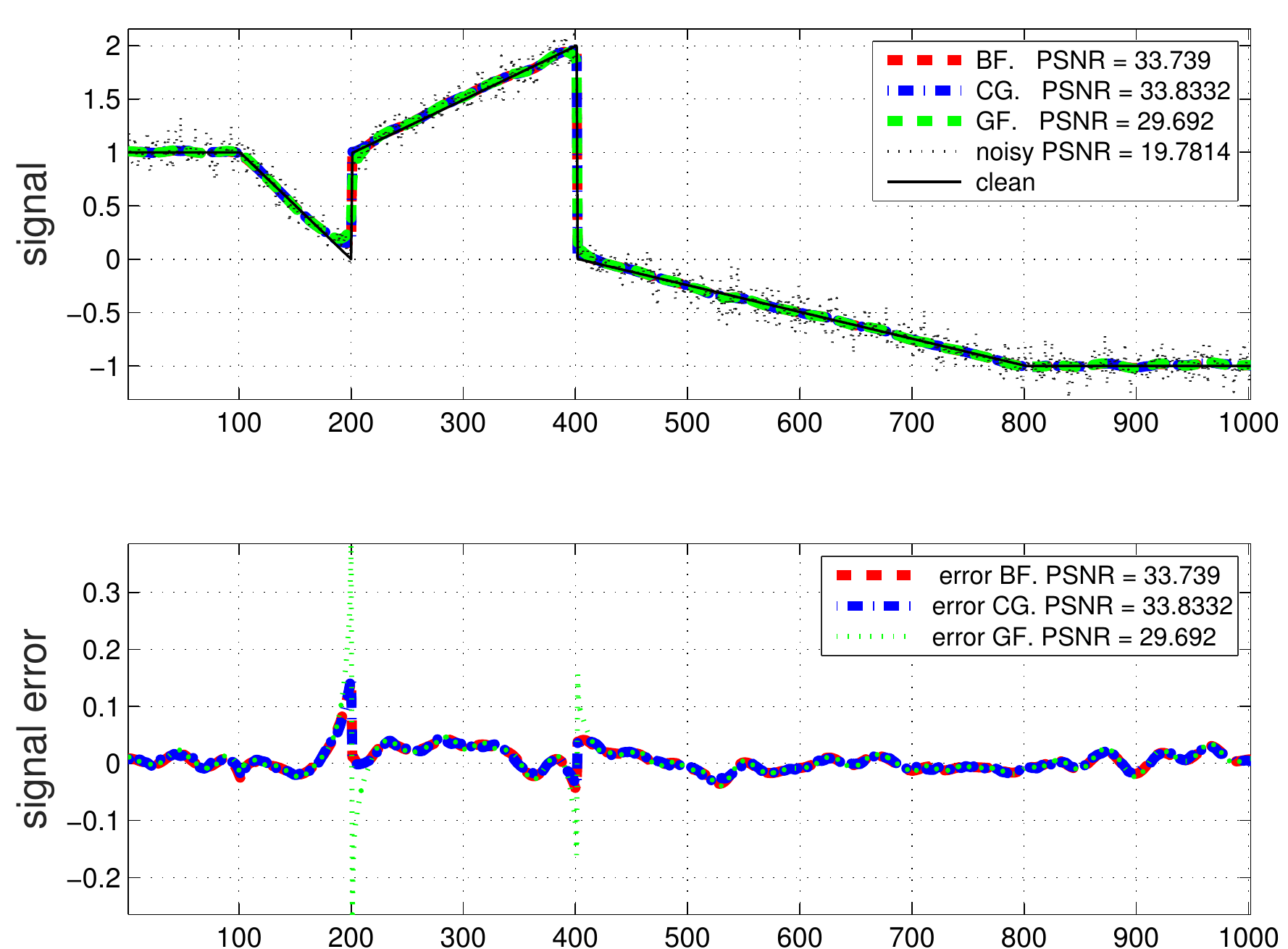}}
\end{minipage}
\caption{Filters in 1D case: 1000 points}
\label{fig:1D1000}
\end{figure}

Our experiments have been done in MATLAB.
We have compared performance of the bilateral filter versus the CG accelerated BF (BF-CG) filter
versus the self-guided GF filter for several one-dimensional signals.
The clean 1D signal is piecewise linear with steep ascents and descents.
The noisy signal is obtained from the clean signal by adding the 
Gaussian noise $0.1\cdot randn(N,1)$, where we test $N=500$ in Figure~\ref{fig:1D500}
and $N=1000$ in Figure~\ref{fig:1D1000}, where
$N$ is number of samples in the signal discretized on a uniform grid.

Figure~\ref{fig:1D500} and Figure~\ref{fig:1D1000} display our test results comparing  
500 iterations of the self-guided BF with the window width equal to 3,
20 iterations of the self-guided GF with the MATLAB default window width 5,
and, finally, 20 iterations of the CG accelerated BF, using the noiseless signal as
a guidance and the diagonal matrix $D$ as the preconditioner.

In Figures~\ref{fig:1D500} and \ref{fig:1D1000}, we observe high quality denoising
by all tested filters, both in terms of the average fit of the noiseless signal
and of preserving the signal details.
We notice that GF has a tendency to round sharp corners compared to BF and
BF-CG, which is not surprising, since GF performs extensive smoothing on
relatively wider neighborhoods. Interestingly, GF error is strongly discontinuous
on the signal edges, while BF and BF-CG provide more conservative approximations.
Increasing the number of samples apparently makes the BF and BF-CG errors
smoother as can be seen comparing Figure~\ref{fig:1D500} and Figure~\ref{fig:1D1000}.

Miraculously all three filters give nearly the same error with the reported parameters,
which requires further investigation and explanation.

Our second series of tests deals with image denoising.
We use the 5-point stencil to generate the lattice for the graph Laplacian matrix
and the edge weights are determined by the BF filter with $\sigma_r=0.1$.

\begin{figure}[htb]
\begin{minipage}[b]{1.0\linewidth}
  \centering
  \centerline{\includegraphics[width=\columnwidth]{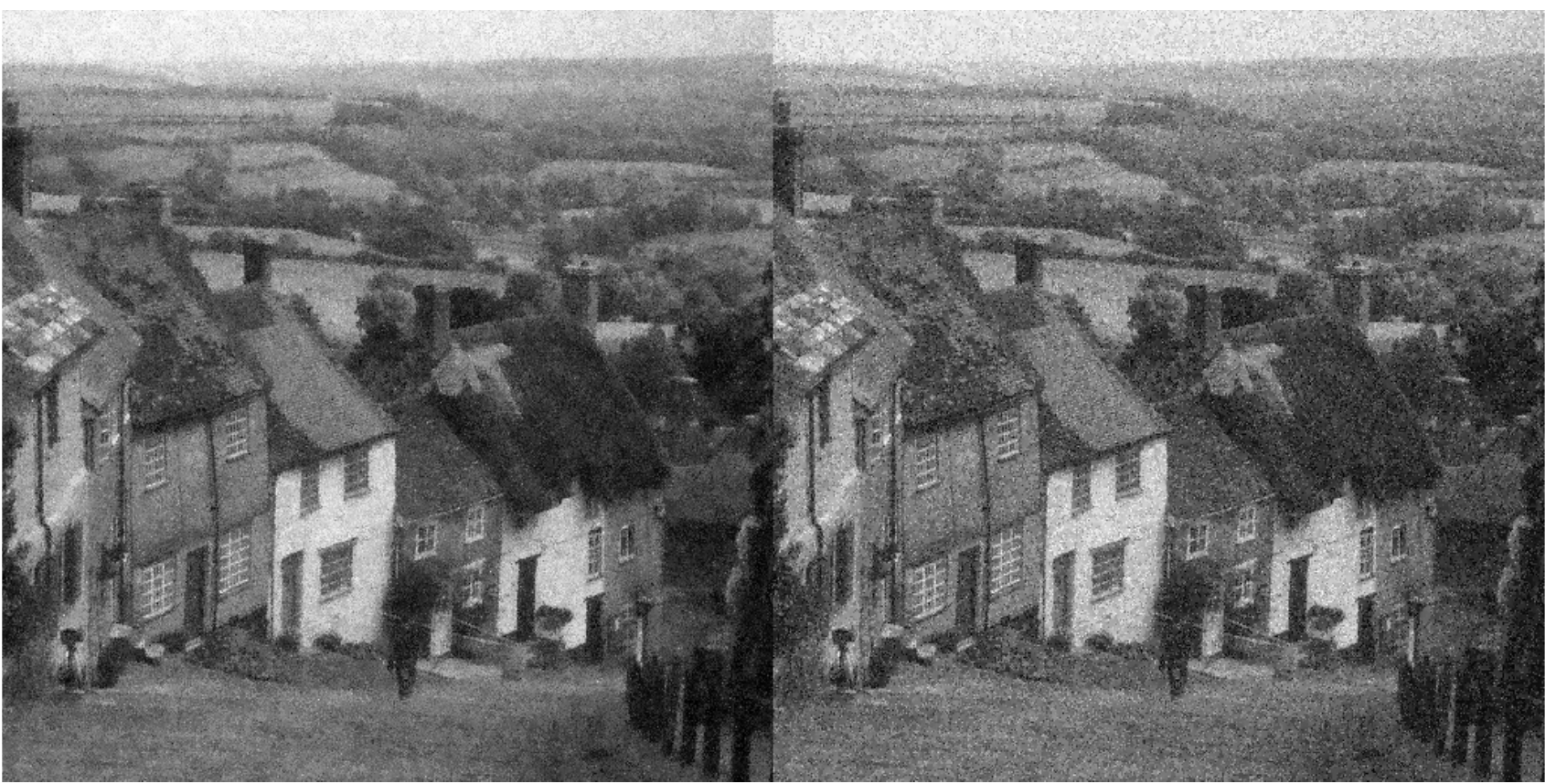}}
\end{minipage}
\caption{BF filter versus CG}
\label{fig:bcomp}
\end{figure}

Twenty iterations are performed using the CG and LOBPCG accelerated filters
with and without the constraint.
The quality of denoising is similar improving PSNR from approximately 20
for the noisy image to approximately 21 for the filtered images.
The edges are well preserved but one can notice forming salt and pepper noise,
which gets more extensive if the number of iterations is increased.
Figure~\ref{fig:bcomp} compares the BF filter with the window half-width 5 and
bilateral filter with the standard deviation $\sigma=0.1$.

\begin{figure}[htb]
\begin{minipage}[b]{1.0\linewidth}
  \centering
  \centerline{\includegraphics[width=\columnwidth]{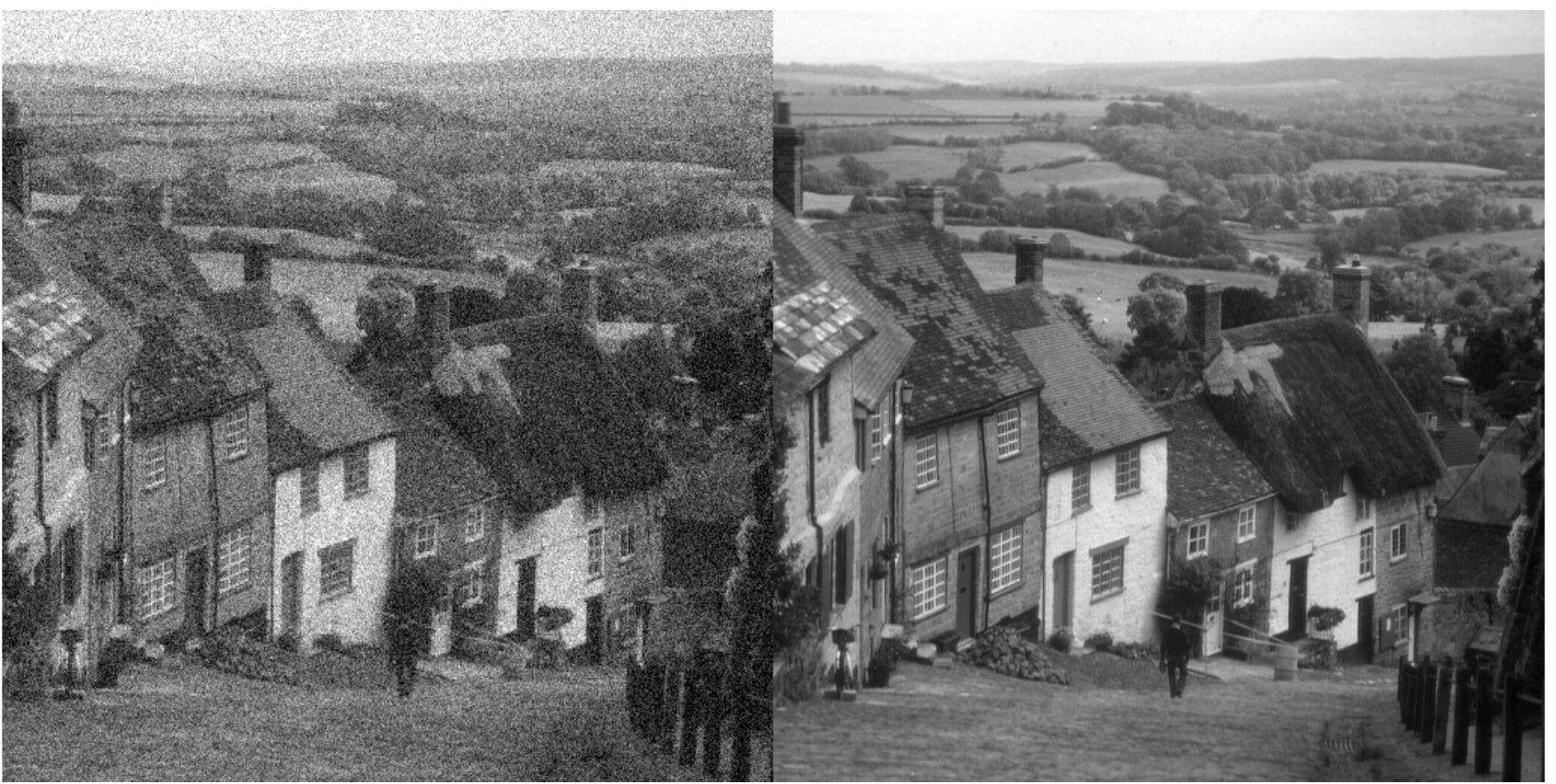}}
\end{minipage}
\caption{Noisy versus clear image: gaussian noise, mean = 0, variance = 0.01, PSNR = 20.1}
\label{fig:ball}
\end{figure}

\begin{figure}[htb]
\begin{minipage}[b]{1.0\linewidth}
  \centering
  \centerline{\includegraphics[width=\columnwidth]{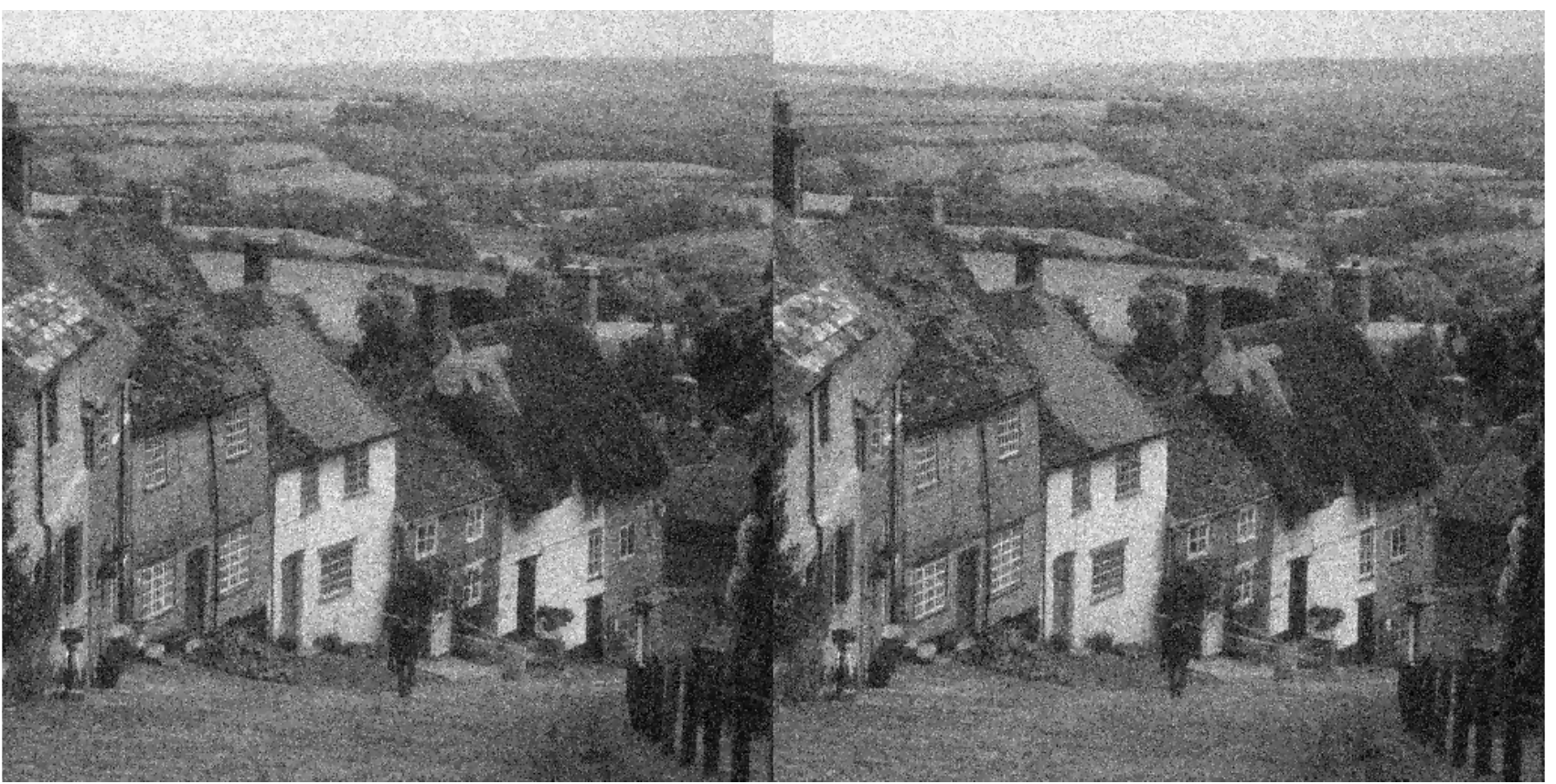}}
\end{minipage}
\caption{LOBPCG with versus without the constraint}
\label{fig:ball1}
\end{figure}

\begin{figure}[htb]
\begin{minipage}[b]{1.0\linewidth}
  \centering
  \centerline{\includegraphics[width=\columnwidth]{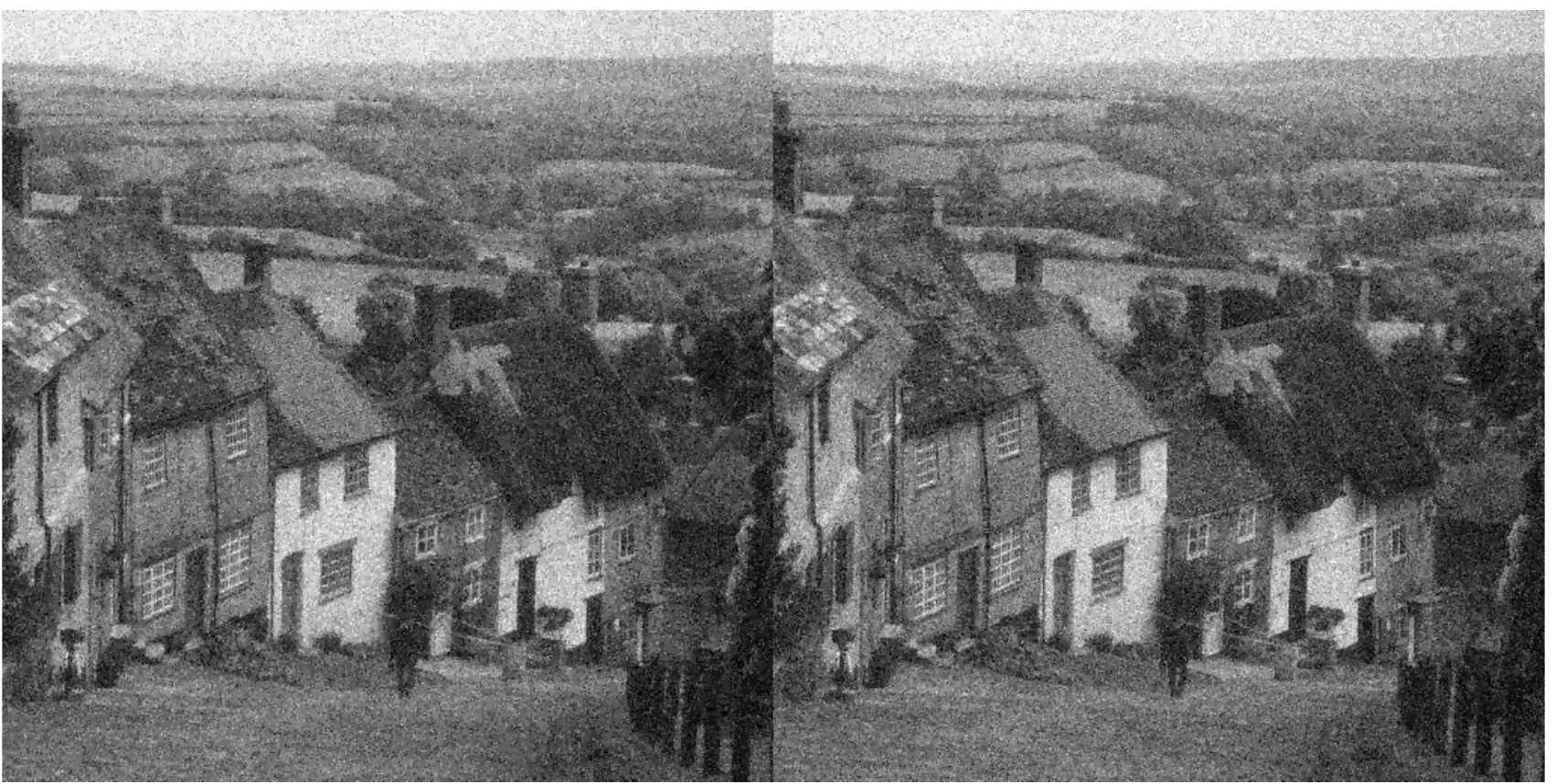}}
\end{minipage}
\caption{LOBPCG with the constraint versus CG}
\label{fig:bcomp1}
\end{figure}

\section*{Conclusions}

We propose accelerating iterative smoothing filters such as the bilateral and guided filters,
by using the polynomials, generated by the conjugate gradient-type iterative solvers
for linear systems and eigenvalue problems.
This results in efficient parameter-free polynomial filters in the Krylov subspaces
that approximate the spectral filters corresponding to the graph Laplacians, which are
constructed using the guidance signals.
Our numerical tests for one-dimensional signals demonstrate the typical behavior of the proposed
accelerated filters and explain the motivations behind the construction.
The tests showing image denoising reveal that the proposed filters are competitive.
Our future work will concern accelerating the guided filters,
testing the influence of non-linearities on the filter behavior,
and incorporating efficient preconditioners. 

% Below is an example of how to insert images. 
% \begin{figure}[htb]
% 
% \begin{minipage}[b]{1.0\linewidth}
%   \centering
%   \centerline{\includegraphics[width=8.5cm]{image1}}
%   \centerline{(a) Result 1}\medskip
% \end{minipage}
% 
% \begin{minipage}[b]{.48\linewidth}
%   \centering
%   \centerline{\includegraphics[width=4.0cm]{image1}}
%   \centerline{(b) Results 3}\medskip
% \end{minipage}
% \hfill
% \begin{minipage}[b]{0.48\linewidth}
%   \centering
%   \centerline{\includegraphics[width=4.0cm]{image1}}
%   \centerline{(c) Result 4}\medskip
% \end{minipage}
% 
% \caption{Example of placing a figure with experimental results.}
% \label{fig:res}
% \end{figure}


% BiBTeX files (here: refs). The IEEEbib.bst bibliography
%\cite{TM98}
\balance
\bibliographystyle{IEEEbib}
\bibliography{refs}
\end{document}